\documentclass{article}
\usepackage{spconf,amsmath,graphicx}
\usepackage{xcolor,hyperref}
\usepackage{caption}

\newcommand{\etal}{\textit{et al.\hspace{0.5mm}}}

\title{Teach me sign: stepwise prompting LLM for sign language production}
\name{Zhaoyi An\thanks{© 2025 IEEE. Personal use of this material is permitted. Permission from IEEE must be obtained for all other uses, in any current or future media, including reprinting/republishing this material for advertising or promotional purposes, creating new collective works, for resale or redistribution to servers or lists, or reuse of any copyrighted component of this work in other works.}, Rei Kawakami}
\address{Institute of Science Tokyo}

\begin{document}
%
\maketitle

%
\begin{abstract}
Large language models, with their strong reasoning ability and rich knowledge, have brought revolution to many tasks of AI, but their impact on sign language generation remains limited due to its complexity and unique rules.
In this paper, we propose TEAch Me Sign (TEAM-Sign), treating sign language as another natural language. By fine-tuning an LLM, we enable it to learn the correspondence between text and sign language, and facilitate generation. Considering the differences between sign and spoken language, we employ a stepwise prompting strategy to extract the inherent sign language knowledge within the LLM, thereby supporting the learning and generation process. Experimental results on How2Sign and Phoenix14T datasets demonstrate that our approach effectively leverages both the sign language knowledge and reasoning capabilities of LLM to align the different distribution and grammatical rules between sign and spoken language.
\end{abstract}
\section{Introduction}
\label{sec:intro}

Given textual input, sign language production aims to produce corresponding sign language videos for the hearing-impaired community to understand. As an important part to facilitate seamless communication with the community, this task
is of great significance in helping them integrate into society, enhancing social inclusion and understanding, and appreciating the culture of sign language and the hearing impaired.

Although the task of translating sign language videos back into text has made some progress recently~\cite{tan2024review,zhou2023gloss}, 
SLP, the task focusing on generating sign videos from input text, remains challenging.
This is due to the complexity of the video generation and the need for generating fine-grained hand movements of sign language. Also, compared to the general video generation field, the available datasets in the sign language domain are relatively smaller in scale and limited in quality, further hindering the development of SLP models.

Previous SLP studies~\cite{saunders2020progressive,huang2022dualsign} typically first convert the input spoken language text into sign language glosses, which can be seen as a sequence-to-sequence problem, and then generate sign language videos based on the resulting glosses. These methods usually require supervision at both the gloss and video levels, and by leveraging additional information, they can ensure a certain level of generation quality. 
However, this also means that the quality of video generation is limited by the result of gloss generation, leading to error accumulation. Additionally, as there is no strict standard for gloss annotation, the inconsistencies and gaps in gloss annotation within the data, further impact the model performance. 
In recent years, more researchers have aimed to directly generate sign language videos from spoken language text~\cite{hwang2023autoregressive,baltatzis2024neural,yin2024t2s} to avoid the issues caused by glosses, achieving some progress. Yet the lack of gloss supervision also make it difficult to align the distributions between sign and spoken language, which in turn affects the model's performance.



With the developments of large language models~\cite{touvron2023llama,brown2020language} and vision language models~\cite{liu2023llava,hu2024bliva}, their strong inference ability and sign language knowledge suggest the possibility of being applied to SLP task.
Actually, some studies have utilized them for human motion generation and video generation tasks.  
Among them, some utilize LLM's large training corpora to generate detailed text descriptions from text input~\cite{athanasiou2023sinc,huang2024free}, while others leverage its reasoning ability to generate structured, machine-readable tokens~\cite{zhang2024motiongpt,lian2024llmgrounded}. These descriptions and tokens are then fed into the generation model and the decoder respectively to create visual outputs. 
As generating text descriptions for subtle sign poses can be challenging, 
yielding specific tokens from the LLM seems to be more promising. However, how to design such tokens to represent sign video remains an issue.
For vision language models, as pointed by~\cite{rahmanzadehgervi2024vision}, current VLMs perform poorly with tasks that require precise spatial information awareness, which is exactly the ability that sign language tasks requires. Thus, we opt to utilize an LLM in this paper. 

To utilize the inherent sign language knowledge and powerful reasoning ability of language-only LLMs for SLP task,
we propose Teach Me Sign (TEAM-Sign), which teams up these two abilities in one pipeline, to generate reasonable sign language pose sequences autoregressively.
Specifically, to address both the information bottleneck of two-stage methods and the generation difficulty of direct methods, we propose to prompt the LLM, to generate auxiliary sequences from its inherent sign language knowledge, which is then provided as additional information. These auxiliary sequences contain the order and key words of the input text, gradually ``teaching'' and ``reminding'' the LLM of the correct and uniform steps when expressing the given text in sign language. Different from previous work, our ``assistance'' comes from the LLM itself, instead of any additional data annotation. 
Acknowledging the modality gap between the sign videos and texts, we convert sign language videos into text-format tokens by the VQ-VAE~\cite{van2017neural}, to enable that the sign video information can be accepted by the language-only LLM. 
Finally, with the auxiliary information and text format sign video tokens, we process datasets to finetune the model and utilize LLM's strong reasoning ability to learn the mappings between texts and sign poses to facilitate generation.

Our contributions can be listed as follows:
\begin{itemize}
    \item We propose TEAM-Sign, which, to the best of our knowledge, is the first practice of utilizing an off-the-shelf LLM for the production task of sign language.
    \item To leverage the inherent sign knowledge of the LLM, we propose using carefully designed prompts to extract sign language rules, which are then applied to identify critical words in the text and assist in generation.
    \item Experimental results show that our proposed TEAM-Sign outperforms competing model by 18.7\% on the Phoenix14T dataset and 20.9\% on the How2Sign dataset, verifying the effectiveness of the approach.
\end{itemize}

\section{Related Work}
\label{sec:related}


\textbf{Sign Language Production} \quad Beside traditional SLP methods~\cite{kipp2011sign,mcdonald2016automated}, deep learning-based ones, as explained above, are basically divided into two types, two-stage approaches and direct approaches~\cite{tan2024review}.

Two-stage approaches mainly rely on additional annotated information, such as gloss, to provide extra supervision during the generation phase. For example, Saunders \etal \cite{saunders2020progressive} proposes Progressive Transformer, which adopts transformer architecture to generate gloss from text, then pose from gloss. 
Huang \etal \cite{huang2022dualsign} propose a semi-supervised two-stage SLP framework that leverages partially gloss-annotated text-pose pairs and monolingual gloss data, 
reducing gloss annotation dependence but still being constrained by the initial and pseudo gloss quality.
Arkushin \etal \cite{arkushin2023ham2pose} replace gloss with HamNoSys \cite{walsh2022changing} and propose the first HamNoSys-to-Pose model, enabling multilingual application due to HamNoSys' cross-linguistic coherence.
However, their method still inevitably requires additional annotations, and when dealing with complex sign language videos, the corresponding HamNoSys sequences tend to be excessively long.

Recently more and more approaches choose to produce sign language videos directly from texts, instead of introducing additional supervision. For example, 
Hwang \etal \cite{hwang2023autoregressive} propose encoding sign language poses as discrete tokens via an encoder and argmax operations, then using a transformer encoder-decoder to generate similar tokens from text and decode them into sign language videos.
However, training a transformer from scratch exacerbates the issue of sign language data scarcity, which affects the learning and generation process.
Later, Baltatzis \etal \cite{baltatzis2024neural} proposes a diffusion-based SLP model, to generate dynamic avatars performing sign language. 
Despite being impressive, it remains questionable whether diffusion models can accurately generate the fine details of hand gestures in sign language. Moreover, the lack of additional information in these direct methods makes the alignment between sign language and natural language more challenging, impacting the quality of generation.


\noindent \textbf{LLMs and VLMs} \quad With its massive training data and parameter scale, the emergence of LLMs has brought significant transformations to deep learning and the broader AI field. Although encoder-only and encoder-decoder~\cite{devlin2018bert,lewis2019bart} architectures of LLMs were developed earlier, in recent years, decoder-only architectures represented by models like the GPT~\cite{brown2020language} and LLaMA~\cite{touvron2023llama} series have demonstrated remarkable performance in generative tasks. 
While possessing strong reasoning abilities, they have also acquired knowledge of sign language from the training corpus.

On the other hand, with the success of LLMs in text generation, some researchers have started exploring Vision Language Models (VLMs) that can handle various modalities such as text, images, and videos. There have also been attempts to use these VLMs for video and human motion generation, as seen in work like~\cite{liu2023llava,hu2024bliva,jiang2024motiongpt}.
However, due to the scarcity of sign language data and the requirement for fine-grained detail generation, directly applying these VLMs to learn and generate sign language yields limited results.

\section{Proposed Method}
\label{sec:method}

\begin{figure*}[!htb]
\centering
\includegraphics[width=\textwidth]{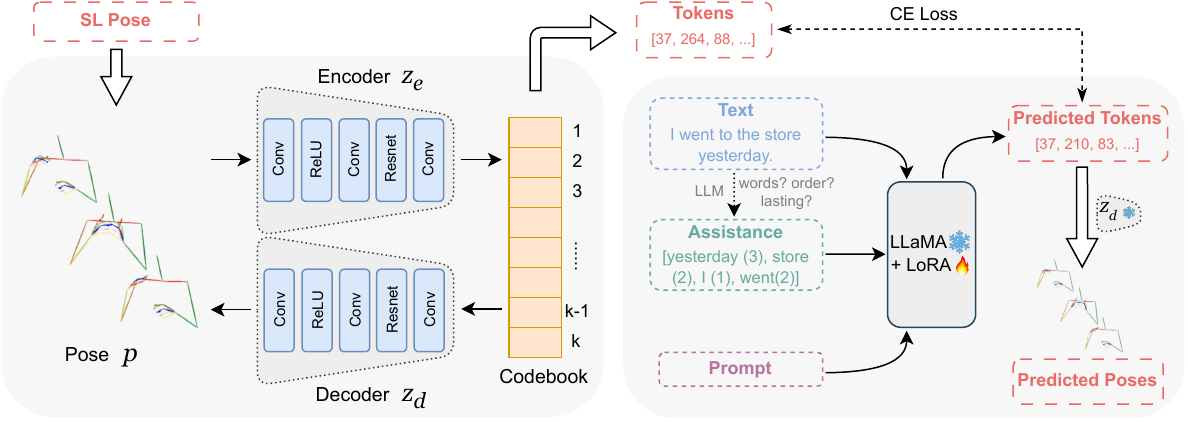} 
\caption{Pipeline of the proposed TEAM-Sign. 
At the sign language encoding stage (left), the sign encoder, decoder and codebook are learned through reconstruction task, in which the sign poses are converted into numerical tokens. For the relation learning stage (right), the text, assistance and prompt are fed into the LLM, to learn the mappings and predict pose tokens. These tokens are then converted back into sign poses through the pose decoder.}
\label{fig:pipeline}
\end{figure*} 

Our proposed TEAM-Sign, as shown in Fig~\ref{fig:pipeline}, leverages the embedded sign language knowledge and reasoning capabilities of LLMs to facilitate the generation from text to sign language video. Through the sign language encoding stage, we address the modality gap between sign language videos and texts by converting the sign video into a sequence of number tokens. However, 
directly learning the relationship between converted sign videos and texts remains challenging. 
To overcome this issue, instead of requiring additional annotations, we opt to utilize the intrinsic sign language knowledge of the LLM itself in the relation learning stage. Specifically, by prompting the LLM to identify keywords and their supposed order in the sign video step-by-step, we provide useful auxiliary information to guide the subsequent learning process, thus enabling the reasoning ability of the LLM to be fully utilized and ensuring generation quality. The generated tokens are then decoded back into videos.

\subsection{Sign Language Encoding}

As previously mentioned, due to the challenges Visual Language Models (VLMs) faced in understanding and generating fine-grained details such as finger movements, we opt to utilize the embedded sign language knowledge and reasoning capability of LLMs for sign language video generation. To enable LLMs, which can only process discrete textual input, to handle sign language video information, we employ the widely-used VQ-VAE~\cite{van2017neural}. This approach maps the features of poses in the sign language video to discrete variables in a codebook and reconstructs the sign language video through these discrete codes.


As shown in Fig.~\ref{fig:pipeline} (Left), in this stage, the input sign language pose \textit{p} is first fed into the pose encoder to generate the embedded feature, which is then compared with each vector in the learnable codebook and quantized to the nearest one, as follows:
\begin{equation}
    z_e(p) \approx e_k, \quad \text{where} \quad k = \arg \min_j \|z_e(p) - e_j\|_2,
\end{equation}
where $e_k$ and $e_j$ are the embeddings from the learnable codebook and $z_e$ is the pose encoder.

In this way, the features of each frame in the sign language video are quantized into feature vectors from the codebook. The corresponding indices of these feature vectors in the codebook are then used to represent the actions in the sign language video. Therefore, an entire sign language video can be represented by a sequence of these indices, denoted as $\{k_1, k_2, ..., k_n\}$. In this stage, we retrieve the corresponding feature vectors from the codebook based on the sequence of indices and attempt to reconstruct the original sign language video using a decoder, which mirrors the encoder's structure. The pose encoder, pose decoder, and codebook are trained through a reconstruction task as
\begin{equation}
    \hat{p} = z_d(e_k).
\end{equation}

As sign language poses always involve with subtle finger movements, in addition to the reconstruction loss, we introduce a new repetition loss during the training process, to prevent their features to be quantized to the same vector in the codebook, and encourage more vectors to be utilized. The VQ-VAE optimization loss is then defined as
\begin{equation}
    L_{Recon} = ||\hat{p} - p||^2 + \alpha ||z_e(p) - sg[e_k]||^2, 
\end{equation}
\begin{equation}
    L_{Repet} = \beta (1 - \frac{k_{uniq}}{len(\{k\})}) + \gamma \frac{k_{freq}}{len(\{k\})}, 
\end{equation}
\begin{equation}
    L_{VQVAE} = L_{Recon} + L_{Repet},
\end{equation}
where sg indicates the stop gradient operation, $\{k\}$ denotes the generated indices from corresponding sign language video, as $\{k\} = \{k_1, k_2, ..., k_n\}$. $k_{uniq}$ and $k_{freq}$ represent the number of unique $k$ values in the sequence and the highest repetition count of any $k$ value in the sequence, respectively. 

\subsection{Relation Learning}

\begin{table*}[!htb]
\centering
\captionsetup{aboveskip=2pt}
\caption{Quantitative results on Phoenix14T dataset.}
\label{tab:phoenix}
\resizebox{0.75\textwidth}{!}{%
\begin{tabular}{l|c|cccc}
\hline
                           & DTW-MJE & BLEU-1 & BLEU-2 & BLEU-3 & BLEU-4 \\ \hline
PT (w/o data augmentation)  & 0.1383 &  9.583  &  3.787  &  1.642  &  0.692  \\
PT                         & 0.1276 &  11.388  &  5.782  &  3.569  &  2.077  \\ \hline
Proposed - LLaMA3 (w/o assistance)   & 0.1204 &  12.861  &  6.105  &  3.719  &  2.636  \\
Proposed - LLaMA3                   & 0.1056 &  \textbf{13.366}  &  \textbf{6.462}  &  \textbf{4.239}  &  \textbf{3.151}  \\ \hline
Proposed - Qwen2 (w/o assistance)   & 0.1051 &  12.950  &  6.129  &  3.865  &  2.813  \\
Proposed - Qwen2                   & \textbf{0.1038} &  13.022  &  6.131  &  4.136  &  3.107  \\ \hline
Groundtruth                & 0.0000 &  30.730  &  20.995  &  15.522  &  12.298  \\ \hline
\end{tabular}%
}
\end{table*}

\begin{table*}[!htb]
\centering
\captionsetup{aboveskip=2pt}
\caption{Quantitative results on How2Sign dataset.}
\label{tab:how2sign}
\resizebox{0.75\textwidth}{!}{%
\begin{tabular}{l|c|cccc}
\hline
                           & DTW-MJE & BLEU-1 & BLEU-2 & BLEU-3 & BLEU-4 \\ \hline
PT (w/o data augmentation)  &  0.1905  &  4.609  &  2.351  &  1.108  &  0.561  \\
PT                         &  0.1733  &  5.490  &  2.744  &  1.262  &  0.628  \\ \hline
Proposed - LLaMA3 (w/o assistance)   &  0.1406  &  9.963  &  5.044  &  \textbf{2.476}  &  1.058  \\
Proposed - LLaMA3                   &  \textbf{0.1371}  &  10.419  &  5.264  &  2.348  &  1.116  \\ \hline
Proposed - Qwen2 (w/o assistance)   & 0.1422 &  9.971  &  5.041  &  2.329  &  1.097  \\
Proposed - Qwen2                   & 0.1397 &  \textbf{10.532}  &  \textbf{5.278}  &  2.415  &  \textbf{1.140}  \\ \hline
Groundtruth                &  0.0000  &  11.230  &  5.469  &  2.500  &  1.228  \\ \hline
\end{tabular}%
}
\end{table*}


After converting the sign language video into discrete vectors from the codebook and representing them with indices, the next step is to learn the mapping between this numerical ``language" and the original text. We employ the LLM, utilizing its logical reasoning capability, to accomplish this task.
Specifically, by prompting the LLM with sentences like ``I went to the barber shop yesterday.'' and ``I ate a cake yesterday.'', we expect it to be able to infer the corresponding sign language indices for ``I'' and ``yesterday'', even for other words.

However, due to differences in word order and emphasis between natural language and sign language, directly learning this mapping poses challenges for the LLM. To address this, we leverage prompt engineering to generate auxiliary sequences using the LLM's inherent knowledge of sign language, as shown in Fig.~\ref{fig:pipeline} (Right). The knowledge has been learned from its original training corpus, and these auxiliary sequences are designed to assist the LLM in learning the mapping, as:

\textit{\textcolor{red}{User:} Introduce \{American/German\} Sign Language sentence structure.}

\textit{\textcolor{red}{Assistant:} (American/German Sign Language sentence structure and rules.)}

\textit{\textcolor{red}{User:} According to the structure, generate sign language sentence for the following text: \{text\}.}

\textit{\textcolor{red}{Assistant:} (Sign language sentence corresponding to the input text.)}

\textit{\textcolor{red}{User:} Estimate the relative duration of each word's motion in a sign video to generate a sequence, while denoting the duration of the shortest word as 1.}

\textit{\textcolor{red}{Assistant:} (Assistance sequence.)}

We prompt GPT-4o~\cite{hurst2024gpt}, due to its superior instruction-following capability, to generate the above auxiliary sequence for each spoken language sentence. The generated auxiliary sequence is then fed into LLaMA 3-8B~\cite{dubey2024llama} and Qwen2-7B~\cite{yang2024qwen2technicalreport} for finetuning, with prompts as below:

\textit{\textcolor{red}{System:} You are a sign language expert. Generate a sequence of number tokens to express the following sentence in \{American/German\} sign language. The assistance indicates the important words, their supposed order and relative duration of each word in its sign video, while the duration of the shortest word is denoted as 1.}

\textit{\textcolor{red}{User:} sentence: \{sentence\} assistance: \{assistance\}.}

During the finetuning stage, the pose encoder, decoder and codebook are frozen. We adopt Low-Rank Adaptation~\cite{hu2021lora} to minimize the discrepancy between the knowledge of LLMs and data of sign language datasets, and optimize LoRA layers' parameters by cross-entropy loss as
\begin{equation}
\begin{aligned}
        \mathcal{L}_\text{LLaMA} &= \text{CE}(\hat{k}, k) \\
                                 &= - \frac{1}{T} \sum_{t=1}^{T} \log P(\hat{k}_t \mid k, \hat{k}_1, \ldots, \hat{k}_{t-1}).
\end{aligned}
\end{equation}

\section{Experiments}
\label{sec:exps}

\begin{figure*}[!htb]
\centering
\includegraphics[width=\textwidth]{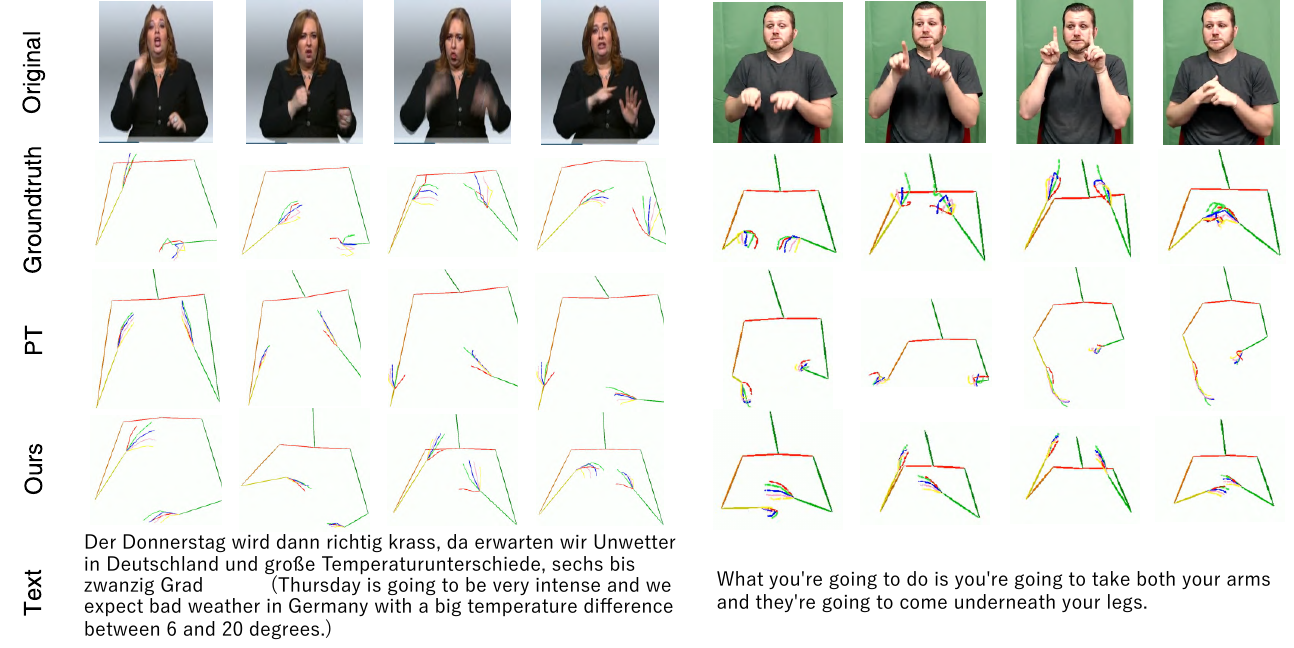} 
\caption{Visual Comparisons on both Phoenix14T (left) and How2Sign (right) datasets.}
\label{fig:subjec}
\end{figure*}

To verify our proposed TEAM-Sign, we conducted comparison experiments on two sign language datasets, namely Phoenix14T~\cite{camgoz2018neural} and How2Sign~\cite{duarte2021how2sign}. As one of the most widely used datasets in the field of sign language recognition and generation, Phoenix14T contains 8,257 German sentences along with their corresponding sign language videos, with a total duration of approximately 10.5 hours. It covers scenarios such as daily news and weather forecasts. On the other hand, How2Sign, an American Sign Language (ASL) dataset, is larger in scale, comprising around 35k English sentences and their corresponding sign language videos, with a total duration of approximately 79 hours. 

\subsection{Quantitative Results}

Following previous practices, we employ DTW-MJE and BLEU$n$ ($n=1,2,3,4$) for objective evaluation. While DTW-MJE computes mean joint error for DTW-aligned pose sequences, BLEU1-BLEU4 originate from the back-translation method proposed in \cite{saunders2020progressive}, used to evaluate the performance of sign language production models. Specifically, the generated video is translated back into spoken language text through another sign language translation model. The BLEU$n$ is calculated as the proportion of $n$ consecutive matching words between the generated text and the reference.

Since pretrained back translation model of previous SLP work is not available, we retrained Zhou \etal \cite{zhou2023gloss} based on keypoint data to get our back translation model for evaluation. For comparison, we adopted ProgressiveTransformer \cite{saunders2020progressive}, to retrain it with and without data augmentation. Furthermore, to investigate the role of the auxiliary sequences generated from the inherent knowledge of the LLM, we also produced a fine-tuning dataset that excludes assistance and contains only the input text and numerical sequences representing sign language videos. 
The LLaMA 3-8B~\cite{dubey2024llama} and Qwen2-7B~\cite{yang2024qwen2technicalreport} are fine-tuned and tested with same settings for comparison.

Quantitative results are tabulated in Table \ref{tab:phoenix} and \ref{tab:how2sign}. From the table, we can see that the proposed model outperforms ProgressiveTransformer \cite{saunders2020progressive} in terms of both DTW-MJE and BLEU$n$ across the two datasets. An interesting finding is that the assistance seems to be more helpful on the smaller Phoenix14T dataset. This could be attributed to the higher complexity of the How2Sign dataset, which makes the differences less noticeable due to the greater challenges in generation and recognition. While another possible reason is that the assistance sentence can supply more missing information on smaller datasets, guiding the model to learn more effectively.

\subsection{Qualitative Results}

Figure~\ref{fig:subjec} displays qualitative comparisons. It can be seen that the keypoint videos generated by our proposed TEAM-Sign are closer to the groundtruth extracted from the original videos in terms of the performance of key actions, such as the motion trends of the arms. On the other hand, the generation of finer details like finger movements still requires further improvement. Notably, due to the limited resolution of some sign videos, issues such as motion blur in fingers also result in inaccuracies in the groundtruth pose obtained through pose estimation (e.g., the 1st and 3rd frames of the ground truth). 
More visual comparisons, a video demo and further discussions are available in the \href{https://drive.google.com/drive/folders/1tdnKoTEEeSjP9y9aBckXKwqOM3cesVWE?usp=sharing}{supplementary material}.

\section{Conclusion}
\label{sec:conclusion}

In this paper, we propose TEAM-Sign, which is, to the best of our knowledge, the first model to utilize LLM's sign language knowledge and reasoning ability simultaneously to perform the sign language production task. Experimental results on the Phoenix14T and How2Sign datasets verify the superiority of our TEAM-Sign method. Additionally, the generation of more fine-grained hand movements and facial expressions remains an open challenge.

This work was supported by JSPS KAKENHI Grant Number JP23H00490 and carried out using the TSUBAME 4.0 supercomputer at Institute of Science Tokyo.



\vfill\pagebreak


\begin{small}
\bibliographystyle{IEEEbib}
\bibliography{strings,refs}
\end{small}

\end{document}